\documentclass{article}

\usepackage[preprint]{neurips_2025}

\usepackage[utf8]{inputenc} 
\usepackage[T1]{fontenc}    
\usepackage{hyperref}       
\usepackage{url}            
\usepackage{booktabs}       
\usepackage{amsfonts}       
\usepackage{amssymb}        
\usepackage{nicefrac}       
\usepackage{microtype}      
\usepackage{graphicx}  

\usepackage{algorithm}
\usepackage{algpseudocode}
\usepackage{float}
\usepackage[table]{xcolor}
\usepackage{amsmath}
\usepackage{colortbl}
\usepackage{tikz}
\usepackage{caption}

\definecolor{mathcolor}{RGB}{235,245,255}
\definecolor{codecolor}{RGB}{245,255,235}
\definecolor{medcolor}{RGB}{255,245,235}
\definecolor{mathcodecolor}{RGB}{240,250,245}
\definecolor{mathmedcolor}{RGB}{245,245,245}

\title{\raisebox{-0.8ex}{\includegraphics[height=0.38in, width=0.38in]{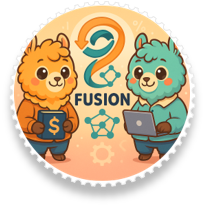}}Graft: Integrating the Domain Knowledge via Efficient Parameter Synergy for MLLMs}
\author{
    Yang Dai\textsuperscript{1} \And 
    Jianxiang An\textsuperscript{1} \And
    Tianwei Lin\textsuperscript{1} \And
    Hongyang He\textsuperscript{1} \And
    Hongzhe Huang\textsuperscript{1} \AND
    Wenqiao Zhang\textsuperscript{1} \And
    Zheqi Lv\textsuperscript{1} \And 
    Siliang Tang\textsuperscript{1} \And
    Yueting Zhuang\textsuperscript{1} \AND
    \vspace{-0.1cm} 
    \textsuperscript{1}\textbf{Zhejiang University} 
    \vspace{0.1cm}
    \\\parbox{0.9\textwidth}{
  \centering
  \texttt{\{yangdai, jianxiangan, tianweilin, hongyanghe, hongzhehuang, wenqiaozhang, zheqilv, siliang, yzhuang\}@zju.edu.cn}
}
}

\begin{document}

\maketitle

\begin{abstract}\phantomsection\label{abs:Abstract}
Multimodal Large Language Models (MLLMs) have achieved success across various domains. However, their applicability tends to degrade when confronted with different types of data inputs, especially for MLLMs that have been fine-tuned for specific tasks. Despite its importance, the study of knowledge sharing among domain-specific MLLMs—such as those trained for mathematics or code—remains largely underexplored.
To address the fragmentation of knowledge across domain-specialized MLLMs, we propose a unified parameter integration framework that enables modular composition of expert capabilities. Our method is grounded in a novel Compatibility-Aware Parameter Splicing (CAPS) strategy, which leverages both local functional attribution and global information-theoretic signals to guide selective parameter fusion. By extending this mechanism to the low-rank adaptation layer granularity, we ensure efficient integration with minimal inference overhead.
Furthermore, we introduce a domain compatibility scoring mechanism that quantifies inter-expert alignment at the activation level and correlates with downstream task utility. This principled fusion protocol allows the final model to synergize heterogeneous expertise while preserving structural modularity.
Extensive evaluations across diverse multimodal benchmarks validate the effectiveness of our framework, offering a scalable path toward compositional, domain-adaptive MLLMs. 


\end{abstract}

\section{Introduction}\label{sec:Introduction}
The development of deep learning is advancing the field of multimodal intelligence and corresponding applications~\cite{li2023winner, zhang2019frame, li2022hero, zhang2022magic, li2022end, zhang2021consensus, zhang2020relational, li2023multi, zhu2023meter, li2023unsupervised}. Recent multimodal intelligence works -
Multimodal large language models (MLLMs) \cite{liu2023visual, wang2024qwen2, liang2024survey}, have emerged as a powerful paradigm in machine learning, have demonstrated remarkable success across various vision-language tasks, such as general reasoning, mathematics, programming, and scientific applications\cite{dyer2022minerva, lin2025healthgpt, liu2025fin, hui2024qwen2,tang2025think}. However, most of them cannot excel in all domains,  mainly due to they were trained on domain-specific settings. Of course, we can introduce more data from different domains and train a comprehensive model from scratch, but it requires significant computational resources. Consequently, there has emerged a recent trend in the research community, \emph{i.e.}, Model Merging\cite{yang2024model, akiba2025evolutionary, li2023deep}, focused on exploring methodologies for effectively merging multiple independently trained models without relying on their training data. The practice of model merging has emerged as a promising solution to enhance model generalization.

Broadly, the existing model merging methods rely on direct integration of model parameters\cite{gupta2020stochastic,wortsman2022model,lv2025optimize}, but these methods presuppose uniform architectures across models and often fail to capture the strengths of diverse specialized models. More advanced heuristics like Task Arithmetic\cite{ilharco2022editing} and TIES-Merging\cite{yadav2023ties}  fuse parameters in an element-wise fashion, but still fail to adequately address parameter interference or to align heterogeneous representations.  
These shortcomings are further exacerbated when merging LoRA-tuned models across disparate domains: misaligned parameter subspaces and an inability to identify which adaptations are complementary versus conflicting often result in severe performance degradation. Collectively, these limitations highlight the need for a principled model fusion strategy capable of adaptively aligning and integrating multi-domain knowledge.



To address these challenges, we propose a novel parameter fusion method named \includegraphics[height=0.25in, width=0.25in]{figure/logo.png}\textbf{Graft}, aiming for more precise and efficient integration of parameters from multiple fine-tuned models. 
The \textbf{GraftModel} variant handles fusion of fully fine-tuned model parameters, while the \textbf{GraftLoRA} variant handles fusion of LoRA-adapted model parameters. 
This dual capability enables flexible knowledge integration from both standard fine-tuned models and LoRA-adapted models. At the local scale, Graft employs a learnable parameter network to measure channel-wise differences, assigning fine-grained weights based on parameter significance. At the global scale, we introduce an entropy-based evaluation mechanism that dynamically adjusts fusion weights according to overall parameter information entropy. By synergistically combining these local and global assessments through a nonlinear adaptive strategy, Graft effectively mitigates the inherent limitations of conventional linear fusion methods.

Moreover, we further ensure fusion performance through an activation-based compatibility analysis method. Specifically, this approach evaluates a model's suitability for fusion by systematically analyzing activation patterns and sensitivities within model modules when mismatched datasets (e.g., mathematics data tested on coding-specific models) are introduced. Such analysis provides crucial insights, significantly improving fusion decision reliability.

Our contributions are summarized as follows: 
(1) We present a novel dual-mode fusion framework that can either merge fully fine-tuned models or LoRA-tuned adapters; 
(2) We develop a comprehensive local-global parameter fusion strategy, enabling precise evaluation and effective integration of diverse model parameters; 
(3) We introduce a learnable parameter network to capture intricate local differences, substantially enhancing fusion accuracy; 
(4) We propose a dynamic entropy-based weighting mechanism, enhancing adaptability and generalization; 
(5) We present a novel single-dataset activation-based compatibility analysis to bolster the reliability of model fusion decisions. 

Collectively, these innovations position Graft as a highly efficient and adaptive parameter fusion method, contributing meaningful theoretical advancements and practical tools that substantially elevate the generalization performance and real-world applicability of large language models.
\begin{figure*}
    \centering
    \includegraphics[width=0.95\textwidth]{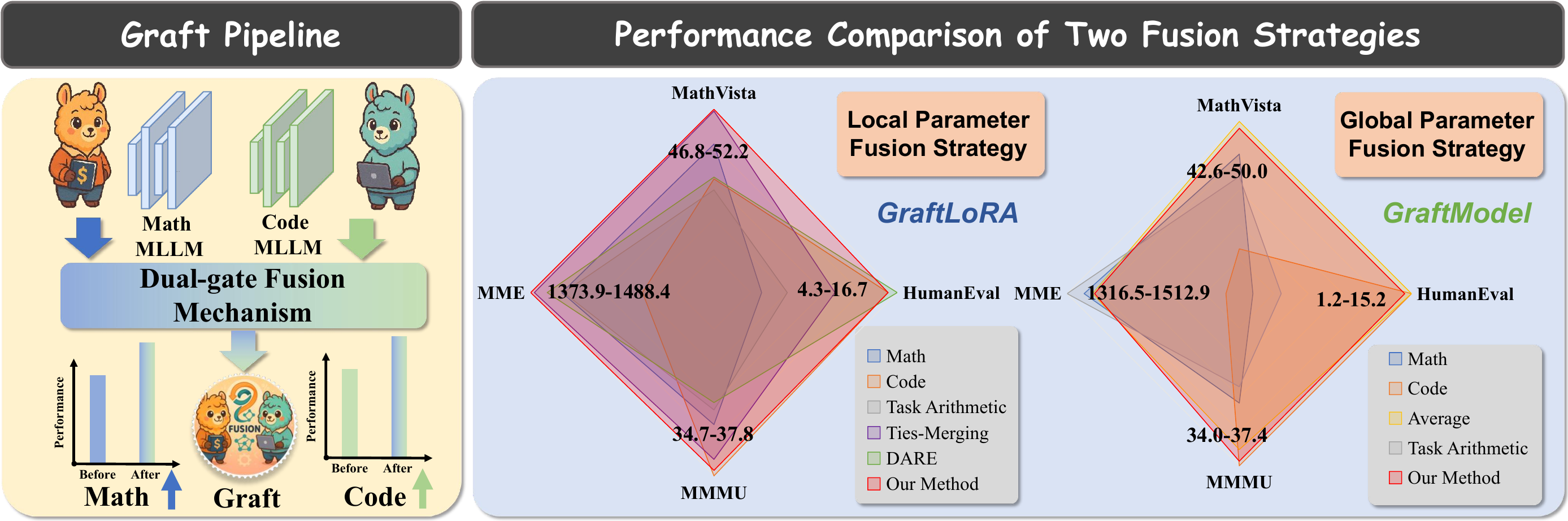}
    \caption{Performance comparison between Graft and other methods on fully fine-tuned and LoRA fine-tuned models.}
    \vspace{-3mm}
    \label{fig:introduction}
\end{figure*}
\section{Related Work}

\noindent\textbf{Foundation Model Fine-tuning.} 
The development of AI has transferred deep learning with small models~\cite{zhong2016overview,zhang2022boostmis,lai2019comparison,zhang2024revisiting,zhang2023learning,lv2023duet,liu2022devrf,ong2022application,lim2022improved,lv2024intelligent,wang2024bridging} to large language models (LLMs). LLMs acquire domain-specific expertise through Supervised Fine-Tuning (SFT), which adapts pre-trained models to excel in targeted domains. To maintain their original, general capabilities while instilling specialized knowledge, a hybrid strategy interleaves a controlled fraction of general-domain data into the fine-tuning corpus \cite{que2024d}. SFT methodologies can be divided into two paradigms based on parameter-update mechanisms: Full Fine-Tuning, which updates all model parameters and is most effective when abundant data and computational resources are available \cite{devlin2019bert,radford2018improving}, and Parameter-Efficient Fine-Tuning (PEFT), which freezes the majority of pre-trained weights and updates only a minimal set of additional parameters, thereby drastically reducing both computational cost and storage requirements \cite{hu2022lora,lester2021power,liu2021p}.

The emergence of domain-specific capabilities in vertically specialized models manifests as a measurable divergence in the weight space—namely, the difference between pre-SFT and post-SFT parameter configurations. This divergence can be formalized as a domain-adaptation vector representation that quantifies the efficiency of task-specific learning \cite{ilharco2022editing}. Despite these advances, integrating multiple vertically specialized models remains an open challenge, as systematic identification, extraction, and fusion of heterogeneous domain vectors are required to achieve synergistic multimodal and multitask performance. Addressing this frontier will demand novel methodologies for disentangling and recombining parameter-space discrepancies across specialized domains.

\noindent\textbf{Model Merge.}
Domain model merging techniques~\cite{} aim to efficiently construct cross-domain generalized models through the integration of model parameters across multiple domains without the need for computationally intensive GPU-based retraining. Early studies\cite{gupta2020stochastic,wortsman2022model} employed parameter averaging strategies that simply computed the arithmetic mean of model weights from multiple domains. Although this approach demonstrated moderate performance improvements in multi-domain tasks, it did not adequately address differences in parameter significance across domains. Subsequent research introduced mechanisms to evaluate parameter importance, such as Fisher Merging 
\cite{matena2022merging}, which leverages the Fisher information matrix to assign weighted factors during parameter updates, and RegMean\cite{jin2022dataless}, which constructs parametric mappings through local regression methods. Nevertheless, these techniques exhibit high computational complexity, limiting their widespread adoption.

More recently, lightweight fusion paradigms have attracted considerable attention. Task Arithmetic\cite{ilharco2022editing}decomposes the fine-tuning process into additive ``task vectors'' represented by the difference between the pre-trained model parameters and fine-tuned parameters, enabling flexible combinations across tasks. Similarly, Ties-Merging\cite{yadav2023ties} alleviates inter-task conflicts through parameter pruning and sign alignment techniques; however, its reliance on global merging coefficients limits fine-grained task-specific adaptability. In contrast, Ada-Merging \cite{yang2023adamerging}introduces a learnable multi-dimensional weighting mechanism, dynamically adjusting parameter contributions through unsupervised optimization. Despite this innovation, the complex training procedure and scenario-specific dependence pose significant challenges for practical deployment.

\noindent\textbf{Multimodal Large Language Models.}
In recent years, the development of deep learning has brought prosperity to the field of multimodal intelligence~\cite{ji2025backpropagation, yao2024mart, yuan2025eoc,jiang2022weakly,zheng2024makima,liu2024boosting,zhang2020photo,yao2023denoising}. Recent progress, Multimodal Large Language Models (MLLMs)have demonstrated remarkable performance in cross-modal tasks~\cite{wu2023multimodal,zhang2024mm,yuan2025videorefer,lin2025healthgpt,zhang2024hyperllava,xie2025heartcare}, such as visual question answering (VQA) and image-text reasoning. Early studies, such as CLIP\cite{radford2021learning}, established a foundation for cross-modal understanding by leveraging contrastive learning to align image and text representations. Meanwhile, generative architectures exemplified by the DALL·E\cite{ramesh2021zero}validated the potential of generative multimodal models. With the maturation of Transformer architectures, multimodal models based on Large Language Models (LLMs), such as LLaVA\cite{liu2023visual}, GPT-4V\cite{yang2023dawn}, CogVLM\cite{wang2024cogvlm}, and Qwen2.5-VL\cite{bai2025qwen2}, have emerged prominently, achieving impressive results in tasks like visual question answering.
However, the high computational costs of training remain a significant obstacle to the widespread deployment of MLLMs. General-purpose multimodal models require extensive resources for pre-training and fine-tuning, while domain-specific variants incur even higher costs. Model merging techniques address this challenge by integrating multiple specialized domain models to efficiently build general-purpose multimodal models, significantly reducing the computational resources required compared to training models from scratch. Particularly promising are fusion approaches applied to homogeneous MLLMs trained across different domains, benefiting from consistent parameter structures.

\begin{figure*}
    \centering
    \includegraphics[width=\textwidth]{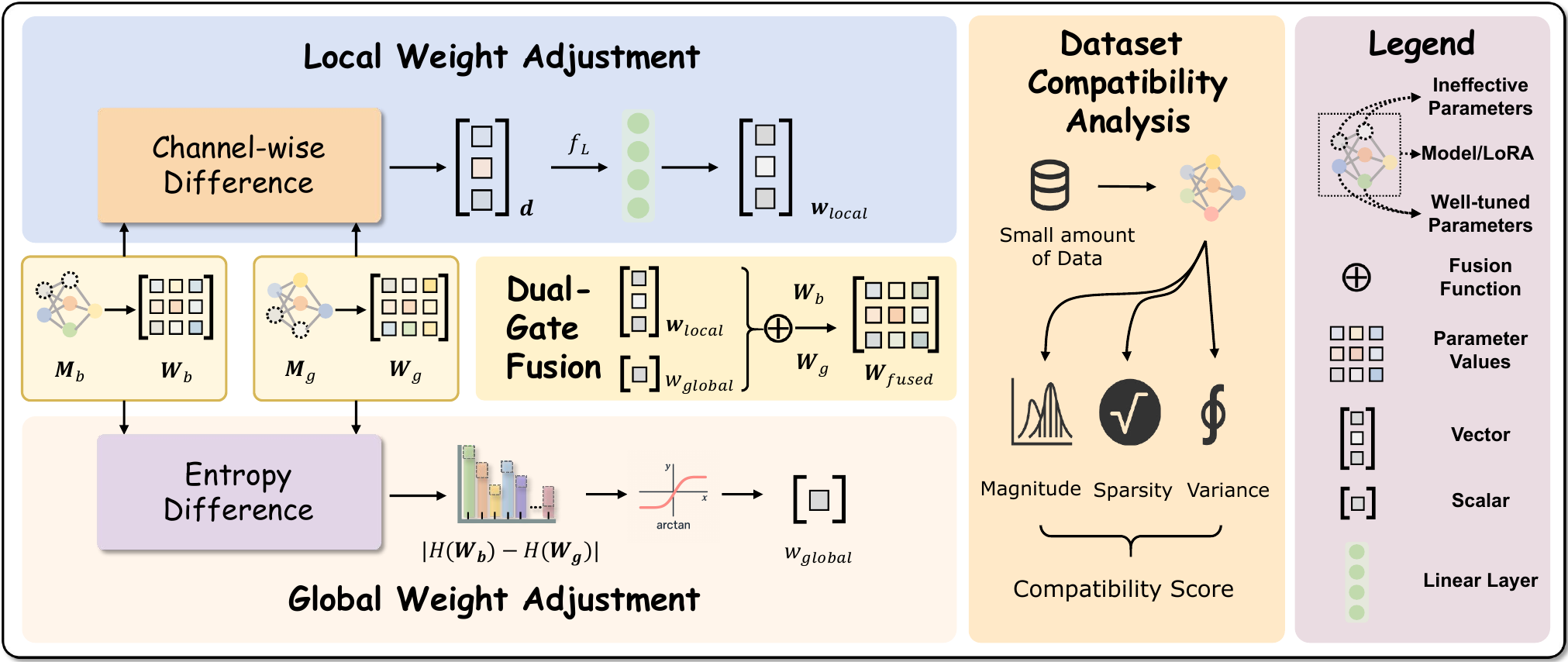}
    \caption{Overview of the proposed Graft method, illustrating how base and graft model or LoRA module parameters are fused via a dual-gate fusion mechanism.}
    \vspace{-3mm}
    \label{fig:method}
\end{figure*}

\section{Methodology}

Our approach aims to integrate two distinct modules—\emph{base} and \emph{graft}—into a unified, parameter-efficient module. To systematically achieve this integration, we propose a dual-gate fusion mechanism that simultaneously leverages \textbf{local, channel-level} discrepancies (Section \ref{method: local}) and \textbf{global, distribution-level} divergences of the parameters (Section \ref{method: global}), thereby enabling adaptive and informed parameter selection. Moreover, to improve the reliability of fusion decisions across different datasets, we further introduce dataset compatibility analysis, which measures a model’s suitability for fusion using an activation-based compatibility metric (Section \ref{method: compatibility}).

\subsection{Local Weight Adjustment} \label{method: local}

To leverage the strengths of each module on a per-feature basis, we first propose a fine-grained \textbf{local weight adjustment} mechanism that dynamically decides, \emph{for each output channel}, whether to emphasize the base module or the graft module. The local weight adjustment mechanism focuses on \emph{channel-wise differences} between these modules. Intuitively, if the two modules differ significantly in a particular output channel, it indicates that they contribute different information for that channel. Therefore, we quantify this difference using the absolute difference between the modules' parameters and use it to guide channel-specific gating decisions.

Let $\mathbf{W}_b \in \mathbb{R}^{M \times N}$ and $\mathbf{W}_g \in \mathbb{R}^{M \times N}$ represent the weight matrices (or flattened parameter sets) of the base and graft modules (e.g., low-rank adaptation layers) for a given layer, where $M$ is the number of output channels (neurons) and $N$ is the number of input features. We measure the absolute difference between $\mathbf{W}_b$ and $\mathbf{W}_g$ for each output channel $i$ as follows:
\begin{align}
d_i=\sum\limits_{j=1}\limits^{N}|\mathbf{W}_b[i,j]-\mathbf{W}_g[i,j]|,\mathbf{d}=(d_1,d_2,...,d_M)^\top\in\mathbb{R}^M,
\end{align}
This yields a difference vector $\mathbf{d} = (d_1, d_2, \ldots, d_M)^\top \in \mathbb{R}^M$, where each element $d_i$ captures the total absolute deviation between the two modules' weights in channel $i$. A larger $d_i$ implies that the base and graft adapters disagree more in the $i$th channel (i.e., one adapter has learned significantly different feature importance for that channel than the other). Next, we feed this difference vector into a learnable \textbf{channel-level gating network}, denoted as $\phi(\cdot)$. The gating network $\phi$ is designed to transform the raw differences $\mathbf{d}$ into an informative gating signal. In practice, $\phi$ could be a small fully-connected module or an affine transformation that processes $\mathbf{d}$ (or each $d_i$ independently) and outputs a corresponding set of gating logits. We then apply a sigmoid activation $\sigma(\cdot)$ to obtain a normalized weight between 0 and 1 for each channel:
\begin{align}\mathbf{w}_{local}=\sigma(\phi(\mathbf{d}))\in (0,1)^M,\end{align}
where $\sigma$ represents the sigmoid activation. This finally produces differentiable channel-wise gating weights $\mathbf{w}_{local}$ that emphasize essential parameters.

\subsection{Global Weight Adjustment} \label{method: global}
We further introduce a \textbf{global weight adjustment} mechanism based on the \emph{overall distribution} of the modules' parameters. By comparing distributional characteristics of $\mathbf{W}_b$ and $\mathbf{W}_g$, this mechanism provides a single scalar gating value, determining which module is generally more informative or confident, guiding the fusion at a macro level.
Our approach uses the concept of \emph{entropy} to quantify the distributional characteristics of each module’s parameters. The entropy of a module's weight distribution indicates the complexity or information content it encodes: higher entropy suggests a more uniform distribution of parameter values, implying richer and more varied information; conversely, lower entropy indicates a more peaked distribution, potentially suggesting sparser or more specialized information. Specifically, we discretize the parameters into $n$ uniform bins to compute the entropy:
\begin{align}H(\mathbf{W}=-\sum\limits_{k=1}\limits^{n}p_k\log p_k,\ p_k=\frac{|w\in\mathbf{W}|w\in B_k|}{M\times N})\end{align}
where the numerator is the number of elements of $\mathbf{W}$ whose value lies in the interval defining bin $B_k$, and $M \times N$ is the total number of parameters in $\mathbf{W}$. Based on the entropy difference between base and graft adapters, we determine a global fusion scalar weight:
\begin{align}w_{global}=\frac{a}{c}\arctan(c[H(\mathbf{W}_b)-H(\mathbf{W}_g)])+\frac{1}{2}\in(0,1),\end{align}
where $a$ and $c$ are constants that shape the $\arctan$ function’s output range and slope. Here, $w_{global}$ is a scalar constrained to $(0,1)$, serving as a global gating factor. In summary, the global weight adjustment encapsulates a high-level judgment of which module appears to carry more information content in its parameters.

\subsection{Dual-Gate Fusion Strategy}
The final fusion incorporates both local and global gating weights to construct comprehensive fusion weights:
\begin{align}\tilde{w}_b=w_{global}(1-e^{-w_{global}\mathbf{w}_{local}}),\end{align}
\begin{align}\tilde{w}_g=(1-w_{global})(1-e^{(1-w_{global})(1-\mathbf{w}_{local})}).\end{align}
These intermediate weights are normalized using softmax to ensure stable and adaptive fusion across all parameter channels:
\begin{align}[w_b,w_g]=\text{Softmax}([\tilde{w}_b,\tilde{w}_g]),\end{align}
\begin{align}\mathbf{W}_{fused}=w_b\odot\mathbf{W}_b+w_g\odot\mathbf{W}_g.\end{align}
This fusion strategy explicitly captures and resolves parameter-level conflicts while optimizing overall model generalization and adaptation capabilities. The overall strategy is summarized as Algorithm \ref{alg:graftlora}:
\begin{algorithm}[H]
\caption{Fusion}
\begin{algorithmic}[1]
\Require Base $W_b\in\mathbb{R}^{M\times N}$, graft $W_g\in\mathbb{R}^{M\times N}$, gate net $\phi$, scalars $a,c$
\Ensure Fused $W_f$
\State $d\leftarrow \sum_j|W_b-W_g|$
\State $D\leftarrow \mathrm{expand}(d)$
\State $w_{\mathrm{loc}}\leftarrow\sigma(\phi(D))$
\State $H_b\leftarrow \mathrm{entropy}(W_b),\;H_g\leftarrow \mathrm{entropy}(W_g)$
\State $w_{\mathrm{glob}}\leftarrow \frac{a}{\pi}\arctan(c(H_b-H_g))+\frac{1}{2}$
\State $\tilde w_b\leftarrow w_{\mathrm{glob}}(1 - e^{-w_{\mathrm{glob}}\,w_{\mathrm{loc}}})$
\State $\tilde w_g\leftarrow (1-w_{\mathrm{glob}})(1 - e^{-(1-w_{\mathrm{glob}})(1-w_{\mathrm{loc}})})$
\State $[w_b,w_g]\leftarrow\mathrm{Softmax}([\tilde w_b,\tilde w_g])$
\State $W_f\leftarrow w_b\odot W_b + w_g\odot W_g$
\end{algorithmic}
\label{alg:graftlora}
\end{algorithm}

\subsection{Dataset Compatibility Analysis} \label{method: compatibility}
In the practice of fusion, selecting appropriate domain-specific models is a crucial step. Models suitable for the target dataset domain can provide a strong starting point for fusion; conversely, mismatched models can even lead to the degradation of the fused model. For more reasonable selection of models, we propose an analysis method to assess dataset compatibility for fully fine-tuned models or LoRA‑adapters fusion at the module level. This analysis introduces an activation-based metric - \emph{compatibility}, indicating the suitability for the given dataset.

Specifically, we choose $K$ input samples from the target dataset, where $K$ is a relatively small value comparing to the total number of samples in the target dataset. Let the activations be $\mathbf{A}_i^{(k)} \in \mathbb{R}^{B \times D}$, where $B$ is the batch size, $D$ the activation dimension, $i$ indexes modules, and $k$ indexes samples. From these activations we compute three statistics per module:
\begin{align}
\text{Mean magnitude}\!: \quad
\mu_i &= \frac{1}{K}\sum_{k=1}^{K}
        \frac{\lVert \mathbf{A}_i^{(k)} \rVert_{1}}
             {\dim\!\bigl(\mathbf{A}_i^{(k)}\bigr)}
        , \\[5pt]
\text{Sparsity}\!: \quad
s_i &= \frac{1}{K}\sum_{k=1}^{K}
      \frac{\#
        \bigl\{\,j : |(\mathbf{A}_i^{(k)})_j| < \epsilon \bigr\}}
           {\dim\!\bigl(\mathbf{A}_i^{(k)}\bigr)}
      , \\[5pt]
\text{Variance}\!: \quad
v_i &= \frac{1}{K}\sum_{k=1}^{K}
      \operatorname{Var}\!\bigl(\mathbf{A}_i^{(k)}\bigr).
\end{align}
Based on these metrics, a comprehensive \emph{data sensitivity} score is computed:

\begin{align}
\rho_i=\mu_i\times(1-s_i)\times\sqrt{v_i}
\end{align}

which quantifies the module's sensitivity to the given dataset. Higher sensitivity scores reflect stronger engagement of the module's parameters, indicating favorable compatibility for fusion.
Moreover, we perform global min-max normalization across modules for each metric, yielding normalized scores $\mu_i^\prime$, $s_i^\prime$, and $v_i^\prime$, enhancing comparability across modules. The normalized sensitivity is then calculated as:
\begin{align}
\rho_i^\prime=\mu_i^\prime\times(1-s_i^\prime)\times\sqrt{v_i^\prime}
\end{align}

Finally, compatibility across all modules is summarized into an aggregate metric:
\begin{align}
\text{compatibility=}\frac{1}{M}\sum\limits_{i=1}\limits^{M}\rho_i^\prime
\end{align}
where $M$ represents the total number of evaluated modules. 
This metric serves as a criterion for evaluating model suitability. In practice, we establish a threshold for this metric. Modules with compatibility exceeding this threshold are considered acceptable for fusion. The utilization of this compatibility metric effectively improves the quality of model fusion and subsequent downstream performance.

\section{Experiments}\label{sec:Experiments}
\subsection{Data and Experimental Setup}
\noindent \textbf{Data Details.} \label{sec:data}
To evaluate the model’s cross‑domain generalization under a controlled data budget, we uniformly sample 5,000 instances from four publicly available corpora: MathV‑360K\cite{shi2024math}, PathVQA\cite{he2020pathvqa30000questionsmedical}, Sujet‑Finance‑QA‑Vision‑100K\cite{SujetFinanceQA2024}, and Code‑Alpaca‑20K\cite{codealpaca}. The first three datasets provide paired image–question–answer triples that span mathematical reasoning, visual pathology diagnosis, and financial chart comprehension, respectively, while Code‑Alpaca‑20K offers purely textual programming instructions. Keeping the sample size constant across all domains eliminates scale‑induced bias and allows us to isolate the effect of modality and semantic diversity on model adaptation.

\noindent \textbf{Experimental Setup.} \label{sec:setup}
We conduct all experiments on the Qwen2‑VL‑2B vision–language model\cite{wang2024qwen2vlenhancingvisionlanguagemodels}. For the hyperparameters in Graft modules, we set the global gating adjustment parameters $a=0.4$ and $c=500$. The entropy calculation uses $n=10$ bins for discretizing weight distributions. All experiments were conducted on 2×A6000 GPUs, using the same hyperparameter settings across all domain adaptation scenarios to ensure fair comparison.
    

\newcommand{\F}{\textsuperscript{F}}  
\newcommand{\Lo}{\textsuperscript{L}} 
\label{tab:main_results}
\begin{table*}[t]
  \caption{Comparison results of model performance (Full vs LoRA) on domain-specific tasks (MathVista, HumanEval) and general benchmarks (MMMU, MME). We use \textbf{bold} text to indicate the best results and {underline} to indicate the second-best results.}
  \label{tab:1}
  \centering
  \setlength{\tabcolsep}{6pt}
  \renewcommand{\arraystretch}{1.1}
  \resizebox{\textwidth}{!}{%
  \begin{tabular}{l|cc|cc||cc|cc}
    \toprule[2pt]
    \multicolumn{1}{c|}{} &
    \multicolumn{4}{c||}{\texttt{GraftModel Performance }} &
    \multicolumn{4}{c}{\texttt{GraftLoRA Performance }} \\
    \cmidrule{2-5}\cmidrule{6-9}
    \textbf{Model} &
    \multicolumn{1}{c}{\textbf{MathVista}\F} & 
    \multicolumn{1}{c|}{\textbf{HumanEval}\F} &
    \multicolumn{1}{c}{\textbf{MMMU}\F} & 
    \multicolumn{1}{c||}{\textbf{MME}\F} &
    \multicolumn{1}{c}{\textbf{MathVista}\Lo} & 
    \multicolumn{1}{c|}{\textbf{HumanEval}\Lo} &
    \multicolumn{1}{c}{\textbf{MMMU}\Lo} & 
    \multicolumn{1}{c}{\textbf{MME}\Lo} \\
    \midrule
    Qwen2‑VL‑2B~\cite{wang2024qwen2} & 47.8 & 14.0 & 34.6 & 1473.5 & 47.8 & 14.0 & 34.6 & 1473.5 \\
    Math                             & 48.1 & 1.2  & 34.7 & \underline{1491.9} & 49.9 & 4.3  & 35.6 & 1455.7 \\
    Code                             & 42.6 & \textbf{15.2} & \textbf{37.4} & 1316.5 & 47.6 & \underline{15.9} & \underline{37.8} & 1373.9 \\
    \midrule
    Average                          & \textbf{50.0} & \textbf{15.2} & 36.7 & 1481.5 & 50.2 & \textbf{16.5} & \textbf{37.9} & 1478.4 \\
    Task Arithmetic\cite{ilharco2022editing}               & 46.8 & 3.7  & 34.0 & \textbf{1512.9} & 46.8 & 6.7  & 35.0 & 1454.0 \\
    Ties‑Merging\cite{yadav2023ties}                        & --   & --   & --   & --     & \underline{52.1} & 11.0 & 37.1 & \underline{1484.1} \\
    DARE\cite{yu2024language}                             & --   & --   & --   & --     & 47.7 & 6.7  & 34.7 & 1471.9 \\
    \textbf{Our Method}              & \underline{49.6} & \underline{14.6} & \underline{37.2} & 1478.9 & \textbf{52.2} & \underline{15.9} & 37.6 & \textbf{1488.4} \\
    \bottomrule[2pt]
  \end{tabular}}
\end{table*}
\subsection{Experimental Results}
\noindent \textbf{Overall Performance Comparison.} 
Table~\ref{tab:1} summarizes the cross‑domain performance of our fusion strategy on four widely used multimodal benchmarks: MathVista, HumanEval, MMMU, and MME. Compared with the pretrained backbone Qwen2‑VL‑2B and four competitive weight‑merging baselines (Average, Task Arithmetic\cite{ilharco2022editing}, Ties‑Merging\cite{yadav2023ties} and DARE\cite{yu2024language}), the proposed dual‑gate Graft delivers the most balanced improvements.\footnote{All methods considered for comparison in this study are fully open‑source; closed‑source or commercial systems are excluded to ensure reproducibility.}

The superscripts F and L in Table~\ref{tab:1} denote fully fine-tuned and LoRA fine-tuned models, respectively. Notably, across all fusion scenarios, the LoRA-tuned domain experts consistently outperform their fully fine-tuned counterparts. For example, fusing LoRA-based adapters yields a MathVista accuracy of 52.2\% compared to 49.6\% with full fine-tuning, and similarly improves the HumanEval pass@1 from 14.6\% to 15.9\%. This trend holds across all evaluated methods, indicating that LoRA preserves complementary knowledge more effectively for model merging. Based on this observation, we conduct all subsequent fusion experiments using LoRA parameters.

\begin{table*}[t]
  \caption{Performance of single‑domain and fused models across multiple domains.
           (\checkmark\ indicates the domain(s) included in the model).}
  \label{tab:domain_fusion}
  \centering
  \setlength{\tabcolsep}{6pt}
  \renewcommand{\arraystretch}{1.1}

  \resizebox{\textwidth}{!}{%
  \begin{tabular}{cccc|cc|cccc}
    \toprule[2pt]
    \multicolumn{4}{c|}{\textbf{Domain Composition}} &
    \multicolumn{2}{c|}{\textbf{Compatibility Scores}} &
    \multicolumn{4}{c}{\textbf{Benchmark Scores}} \\
    \cmidrule{1-4}\cmidrule{5-6}\cmidrule{7-10}
    \textbf{Math} & \textbf{Code} & \textbf{Fin.} & \textbf{Med.} &
    \textbf{Math} & \textbf{Code} &
    \textbf{MathVista} & \textbf{HumanEval} & \textbf{MMMU} & \textbf{MME} \\
    \midrule

    \checkmark &            &            &            & 0.331 & --    & 49.9 &  4.3 & 35.6 & 1455.7\\
               & \checkmark &            &            & --    & 0.286 & 47.6 & 15.9 & 37.8 & 1373.9\\
               &            & \checkmark &            & --    & --    & 43.8 &  8.5 & 36.7 & 1414.7\\
               &            &            & \checkmark & --    & --    & 46.0 & 12.2 & 37.5 & 1470.2\\
    \midrule
    \checkmark & \checkmark &            &            & 0.282 & 0.204 & 52.2 & 15.9 & 37.6 & 1488.4\\
    \checkmark &            & \checkmark &            & 0.280 & --    & 50.1 &  --  & 37.3 & 1470.1\\
    \checkmark &            &            & \checkmark & 0.315 & --    & \textbf{52.4} & -- & 37.0 & \textbf{1535.0}\\
               & \checkmark & \checkmark &            & --    & 0.182 & --   & \textbf{16.5} & \textbf{38.1} & 1457.5\\
               & \checkmark &            & \checkmark & --    & 0.155 & --   & \textbf{16.5} & 38.0 & 1468.6\\
    \bottomrule[2pt]
  \end{tabular}}%
\end{table*}
\noindent \textbf{Cross-Domain Compatibility Analysis.} 
Table~\ref{tab:domain_fusion} extends the compatibility-sensitive fusion analysis beyond the Math–Code pair reported in Table~\ref{tab:1} by evaluating additional cross‑domain settings. Across all benchmarks, the proposed activation‑guided fusion policy consistently outperforms its single‑domain baselines, confirming its domain‑agnostic efficacy. Specifically, Math+Medical attains the highest MME score of 1535.0, representing a 4.4\% relative improvement over the standalone Medical model (1470.2). The Code+Finance fusion yields the best HumanEval accuracy (16.5, +0.6pp), while Code+Medical secures the strongest MMMU result (38.1, +0.3pp). Importantly, these gains arise without additional fine‑tuning, indicating that the compatibility estimator reliably identifies complementary knowledge across heterogeneous domains and thereby provides a plug‑and‑play mechanism for constructing versatile multimodal experts.

The analysis of the compatibility scores based on activation in Table~\ref{tab:domain_fusion} further substantiates their predictive value for the fusion of domains. For Math centric pairs, the higher scores - Math + Medical (0.314) > Math + Code (0.282) $\approx$ Math + Finance (0.280) - align with the larger relative improvements in their primary benchmarks (+4.4\%, +2.3\% and +1.9\% in MME, HumanEval and MMMU, respectively). The Spearman correlation between the compatibility score and absolute performance gain reaches $\rho$ = 0.86, indicating a strong monotonic relationship. An apparent outlier arises in the Code + Medical case: despite a modest score (0.265), the fused model still excels on HumanEval. This behaviour is attributable to the Medical expert’s already competitive baseline on that task, which narrows the observable gain. Consequently, the compatibility score is most informative when interpreted in conjunction with each candidate’s baseline proficiency. We therefore recommend a two‑factor decision rule that weighs (i) the activation compatibility score and (ii) the stronger expert’s standalone performance on the target benchmark to maximise the efficacy of future fusion selections.

\begin{table*}[t]
  \caption{Results of MathVista benchmark across tasks and domains.}
  \label{tab:rotated_mathvista}
  \centering
  \renewcommand{\arraystretch}{1.2}
  \setlength{\tabcolsep}{6pt}
  \resizebox{\textwidth}{!}{
  \begin{tabular}{l|cccccccccccc|c}
    \toprule[2pt]
    \textbf{Domain} & SR & TQA & NC & AR & VQA & GR & ALR & GPS & MWP & LR & FQA & SRG & Overall \\
    \midrule
    \rowcolor{mathcolor}
    Math & 48.4 & 50.0 & 30.6 & 44.5 & 40.8 & 56.5 & 55.2 & 58.2 & 53.8 & 5.4 & 46.8 & 53.5 & 49.9 \\
    \rowcolor{codecolor}
    Code & 55.7 & 51.3 & 31.3 & 44.5 & 51.4 & 35.6 & 37.0 & 34.6 & 41.4 & 8.1 & 57.2 & 58.8 & 47.6 \\
    \rowcolor{mathcodecolor}
    Math\&Code & 58.2 & 51.3 & 33.3 & 50.1 & 49.2 & 47.3 & 44.5 & 46.6 & 54.8 & 13.5 & 57.2 & 61.8 & 52.2 \\
    \rowcolor{medcolor}
    Medical & 53.3 & 47.5 & 31.9 & 41.4 & 48.6 & 36.0 & 35.9 & 35.1 & 38.2 & 13.5 & 57.2 & 56.8 & 46.0 \\
    \rowcolor{mathmedcolor}
    Math\&Med & 54.1 & 48.7 & 34.0 & 50.1 & 49.7 & 50.6 & 47.3 & 50.5 & 53.8 & 10.8 & 56.9 & 61.5 & 52.4 \\
    \bottomrule[2pt]
  \end{tabular}
  }
\end{table*}

\noindent \textbf{Subtask-Level Evaluation.} 
On the twelve sub‑tasks of the MathVista benchmark in Table~\ref{tab:rotated_mathvista}—including Scientific Reasoning (SR), Textbook Question Answering (TQA), Numeric Commonsense (NC), Arithmetic Reasoning (AR), Visual Question Answering (VQA), Geometry Reasoning (GR), Algebraic Reasoning (ALR), Geometry Problem Solving (GPS), Math Word Problem (MWP), Logical Reasoning (LR), Figure Question Answering (FQA), and Statistical Reasoning (SRG)—our fusion models demonstrate a consistent cross‑task advantage.
Taking the Math + Code configuration as an example, the model surpasses the strongest single‑domain baseline by +5.6 points on Arithmetic Reasoning and +2.0 points on Numeric Commonsense, highlighting the complementarity between mathematical representations and programming semantics. In the interdisciplinary MMMU evaluation, the Code + Finance and Code + Medical fusions record the highest scores in Art \& Design (54.6 vs. 53.7), Health \& Medicine (39.6 vs. 38.6), and Humanities \& Social Science (55.0 vs. 54.7). Although the absolute gains over the respective single‑domain models range from 0.3 to 1.6 points, the fused models maintain non‑degraded performance on high‑variance, low‑sample subsets such as Business and Science. These observations corroborate the effectiveness of the entropy‑regularised global gating mechanism in alleviating domain conflict while preserving specialised knowledge during cross‑domain integration.




    




    
\begin{table*}[t]
  \centering
  
  \begin{minipage}[t]{0.48\textwidth}
    \centering
    \captionof{table}{Results of multi‑domain fusion (\checkmark{} indicates included domain).}
    \label{tab:multi_domain}

    \setlength{\tabcolsep}{4pt}
    \renewcommand{\arraystretch}{1.1}
    \resizebox{\textwidth}{!}{
    \begin{tabular}{cccc|cc}
      \toprule[2pt]
      \multicolumn{4}{c|}{\textbf{Domain Composition}} &
      \multicolumn{2}{c}{\textbf{Benchmark Scores}} \\[-1pt]
      \textbf{Math} & \textbf{Code} & \textbf{Finance} &
      \textbf{Medical} & \textbf{MathVista} & \textbf{HumanEval}\\
      \midrule
      \checkmark & \checkmark & \checkmark &            & 51.7 & 14.6 \\
      \checkmark & \checkmark &            & \checkmark & 52.9 & 14.6 \\
      \checkmark & \checkmark & \checkmark & \checkmark & 53.0 & 14.6 \\
      \bottomrule[2pt]
    \end{tabular}
    }
  \end{minipage}
  \hfill
  \begin{minipage}[t]{0.48\textwidth}
    \centering
    \captionof{table}{Ablation study on gating components (\checkmark{} indicates enabled part).}
    \label{tab:ablation}

    \setlength{\tabcolsep}{4pt}
    \renewcommand{\arraystretch}{1.1}
    \resizebox{\textwidth}{!}{
    \begin{tabular}{cc|cccc}
      \toprule[2pt]
      \multicolumn{2}{c|}{\textbf{}} &
      \multicolumn{4}{c}{\textbf{Benchmark Scores}} \\[-1pt]
      \textbf{Local} & \textbf{Global} &
      \textbf{MathVista} & \textbf{HumanEval} & \textbf{MMMU} & \textbf{MME}\\
      \midrule
      \checkmark &            & 52.0 & 15.9 & 37.6 & 1495.0 \\
                 & \checkmark & 51.7 & 12.2 & 37.6 & 1483.4 \\
      \checkmark & \checkmark & 52.2 & 15.9 & 37.6 & 1488.4 \\
      \bottomrule[2pt]
    \end{tabular}
    }
  \end{minipage}
\end{table*}

\noindent \textbf{Multi-Domain Fusion.} 
We next evaluate the scalability of Graft to multi-domain integration by fusing three and four expert adapters. Table~\ref{tab:multi_domain} summarizes results on the two most challenging benchmarks—MathVista and HumanEval. Adding each new expert yields diminishing yet still positive gains on MathVista: fusing Math + Code with the Finance adapter results in an accuracy of 51.7, while substituting Medical further boosts it to 52.9. Integrating all four domains reaches 53.0, delivering a 0.6-point absolute improvement over the best two-domain model. These monotonic gains indicate that heterogeneous domain knowledge compounds to benefit mathematical reasoning.

Coding performance, measured by HumanEval pass@1, remains nearly constant (14.6) as additional domains are grafted. Although the four-domain model falls slightly short of the two-domain peak (15.9), the negligible drop confirms that dual-gating effectively suppresses interference from unrelated experts, preserving the base model’s coding competence. Collectively, these findings demonstrate that Graft scales gracefully beyond pairwise fusion, unifying multiple specialized adapters without catastrophic forgetting. The ability of triple- and quadruple-domain configurations to improve MathVista while maintaining HumanEval underscores the framework’s promise for constructing broadly capable multimodal large language models.

\noindent \textbf{Ablation Study.} 
Table~\ref{tab:ablation} compares three gating schemes—Local‑Gate, Global‑Gate, and Dual‑Gate—across four benchmarks. Dual‑Gate consistently outperforms its single‑gate counterparts, achieving 52.2 on MathVista (vs. 52.0 for Local‑Gate and 51.7 for Global‑Gate), 15.9 on HumanEval (matching Local‑Gate and substantially surpassing Global‑Gate’s 12.2), 37.6 on MMMU (on par with both single‑gate variants), and a 1488.4 composite score on MME, the overall best among all settings.

Mechanistically, Local‑Gate learns a channel‑wise importance mask for each LoRA adapter, thereby amplifying fine‑grained, domain‑specific signals. Global‑Gate, in contrast, derives a single fusion weight from the entropy gap of each adapter’s weight distribution, balancing cross‑domain knowledge at a coarse level. Dual‑Gate synergistically combines these perspectives: the local gate preserves salient micro‑features while the global gate, regularized by entropy, mitigates inter‑domain conflicts. This complementary interaction enables the model to retain specialized expertise without sacrificing holistic performance, which explains the superior results observed on all metrics in Table~\ref{tab:ablation}.

\begin{figure*}[t]
  \centering
  
  \begin{minipage}[t]{0.55\textwidth}
    \centering
    \includegraphics[width=\textwidth]{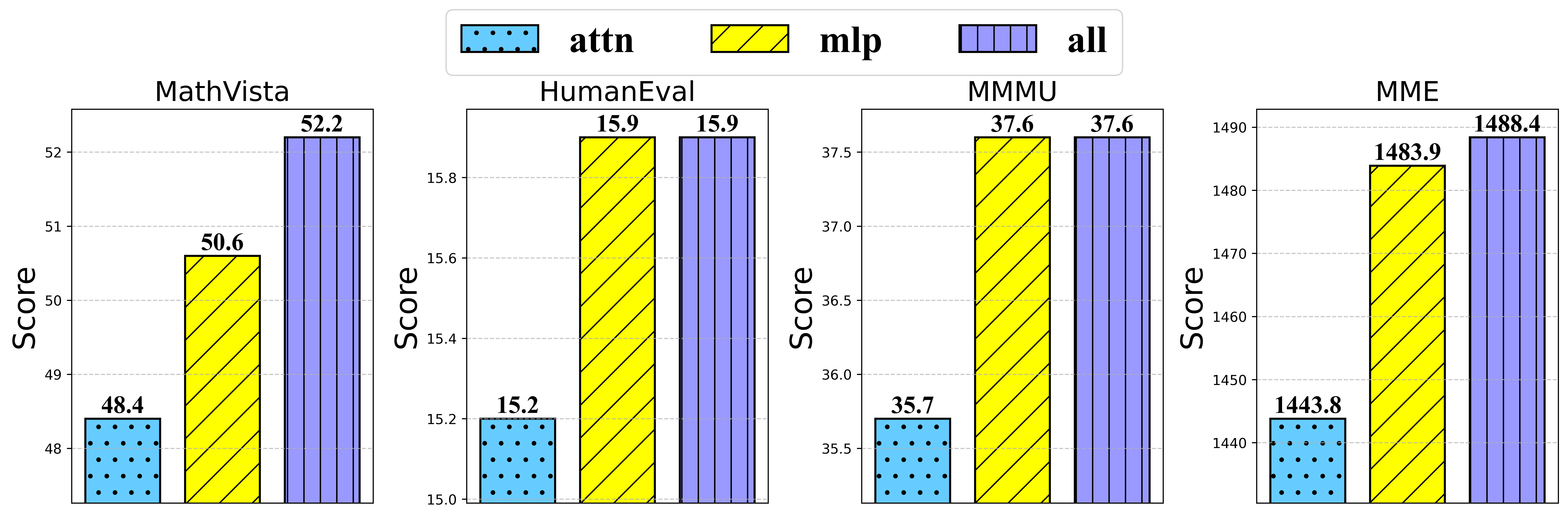}
    \caption{Performance of different projection layer fusion strategies.}
    \label{fig:layer}
  \end{minipage}
  \hfill
  \begin{minipage}[t]{0.4\textwidth}
    \centering
    \includegraphics[width=\textwidth]{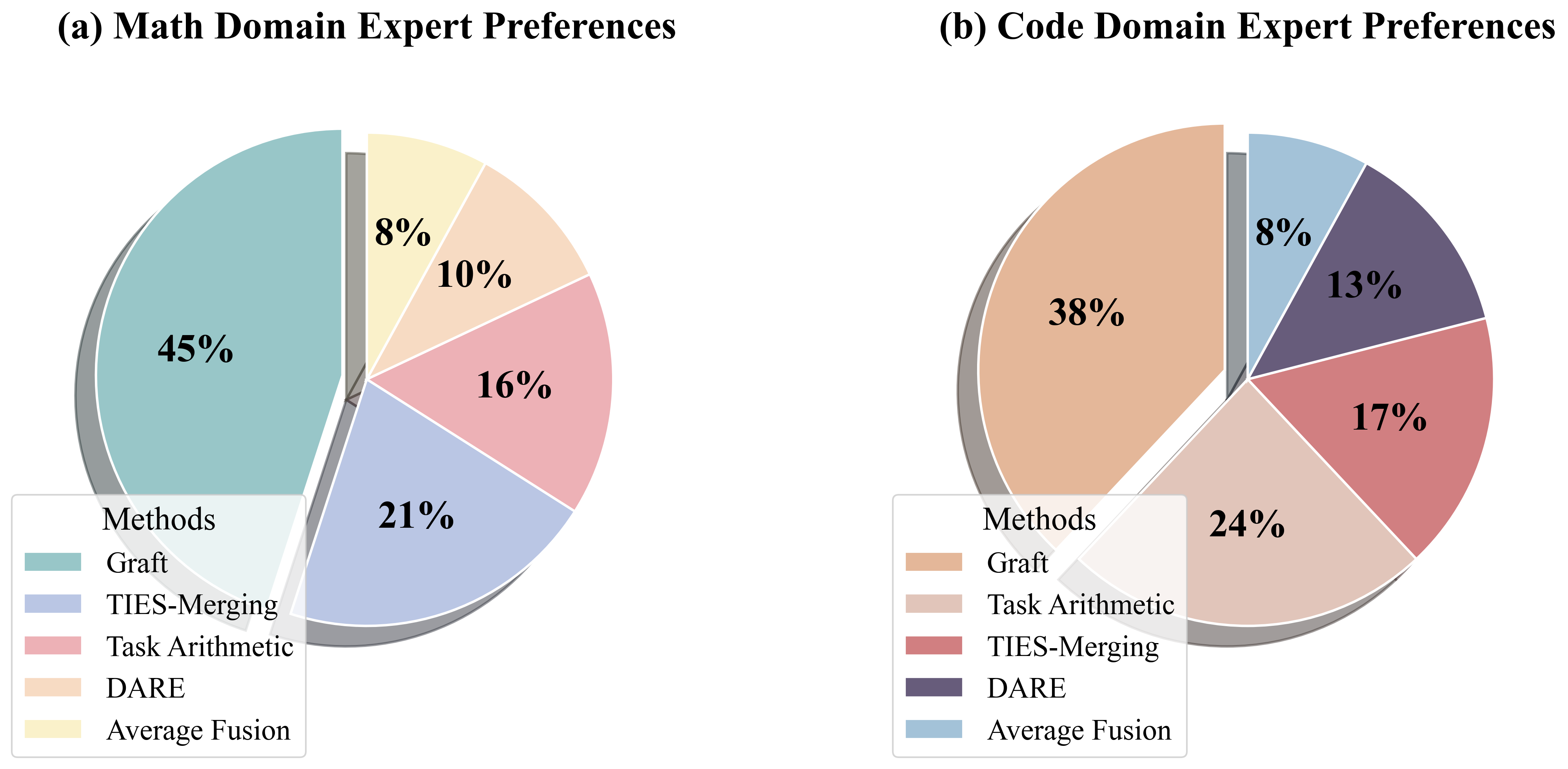}
    \caption{Human preference for generated content of baselines and our model.}
    \label{fig:pie}
  \end{minipage}
\end{figure*}

\noindent \textbf{Layer-wise Fusion Analysis.} 
To investigate the impact of parameter fusion granularity, we conduct studies on selectively merging different projection layers in Transformer blocks. As shown in Figure~\ref{fig:layer}, we compare three fusion strategies: (1) attn: merging only attention projections; (2) mlp: merging only MLP projections; (3) all: jointly merging both attention and MLP projections.

Specifically, merging all projection layers attains 52.2 on MathVista (+3.8 over "attn", +1.6 over "mlp") and 15.9 on HumanEval (+0.7 over "attn"), indicating synergistic benefits from cross-module knowledge integration.The results demonstrate that comprehensive layer fusion achieves optimal performance across all benchmarks, and it validates our design choice of full-layer fusion, which maximizes the preservation of both structural relationships (via attention projections) and feature representations (via MLP projections).

\noindent \textbf{Human Evaluation.} 
We further conduct an expert preference study to evaluate the effectiveness of Graft across domains. We recruited ten domain experts (5 mathematics, 5 computer science) to rank the responses from the fused models of five fusion methods (Average, Task Arithmetic, TIES-Merging, DARE, and Graft) on randomly sampled queries from MathVista and HumanEval datasets. As shown in Figure~\ref{fig:pie}, results demonstrate a clear preference for Graft across both domains. The expert preference results align with our quantitative performances, demonstrating that our dual-gate fusion approach successfully preserves domain-specific knowledge while enabling cross-domain integration.
\section{Conclusion}

In this work, we introduce Graft, a dual-gate parameter fusion framework that synergistically combines local channel-level gating with a global entropy-based weighting mechanism to integrate model parameters from different domain experts. To ensure reliable fusion decisions, we further develop a single-dataset activation-based compatibility analysis that quantitatively predicts complementary domain pairs prior to weight merging.

\bibliographystyle{unsrtnat}
\bibliography{ref}

\end{document}


\appendix
\begin{center}
\Large \textbf{Appendix}
\end{center}
This is the Appendix for ``Graft: Integrating the Domain Knowledge via Efficient Parameter Synergy for MLLMs''. Table~\ref{tab:abb} summarizes the abbreviations and symbols used in the main paper.
\begin{table}[h]
\begin{center}
\captionsetup{font={small,stretch=1.25}, labelfont={bf}}
\caption{Abbreviations and symbols used in the main paper.}
 \renewcommand{\arraystretch}{1.2}
\resizebox{0.48\textwidth}{!}
 {
  \begin{tabular}{c||c }
   \toprule[1.5pt]
   \textbf{Abbreviation/Symbol} & \textbf{Meaning}\\
   \hline
   \hline
   & \underline{\emph{Abbreviation}}\\
   LLMs&Large Language Models\\
   MLLMs&Multimodal Large Language Models\\
   CAPS&Compatibility-Aware Parameter Splicing\\
   SFT&Supervised Fine-Tuning\\
   PEFT&Parameter-Efficient Fine-Tuning\\
   SR&Scientific Reasoning\\
   TQA&Textbook Question Answering\\
   NC&Numeric Commonsense\\
   AR&Arithmetic Reasoning\\
   VQA&Visual Question Answering\\
   GR&Geometry Reasoning\\
   ALR&Algebraic Reasoning\\
   GPS&Geometry Problem Solving\\
   MWP&Math Word Problem\\
   LR&Logical Reasoning\\
   FQA&Figure Question Answering\\
   SRG&Statistical Reasoning\\
   MLP & Multi-Layer Perception\\
     \hline
   \hline
   
      & \underline{\emph{Symbol in Algorithm}}\\   
      $\mathbf{W}_b$ & Base model weight matrix\\
      $\mathbf{W}_g$ & Graft model weight matrix\\
      $\mathbf{d}$ & Channel-wise absolute difference vector\\
      $\phi$ & Learnable channel-level gating network\\
      $\sigma$ & Sigmoid activation function\\
      $\mathbf{w}_{local}$ & Local gating weights from $\phi$ network\\
      $H(\mathbf{W})$ & Entropy function for weight matrices\\
      $H_b,H_g$ & Entropy of $\mathbf{W}_b$ and $\mathbf{W}_g$\\
      $a,c$ & Hyperparameters for global weight adjustment\\
      $w_{global}$ & Global gating scalar from entropy\\
      $\tilde{w}_b, \tilde{w}_g$ & Intermediate fusion weights\\
      $w_b, w_g$ & Normalized fusion weights via softmax\\
      $\mathbf{W}_{fused}$ & Final fused weight matrix\\
      $\mu$ & Mean magnitude\\
      $s$ & Sparsity\\
      $v$ & Variance\\
      $K$ & Numbers of input samples\\
      $\mathbf{A}_i^{(k)}$ & Activations\\
      $\rho$ & Data sensitivity score\\
      $\mu^\prime, s^\prime, v^\prime\, k^\prime, \rho^\prime$ & Normalized scores\\

   \toprule[1.5pt]
  \end{tabular}
  }
  \label{tab:abb}
\end{center}
\end{table}
This Appendix is organized as follows: 
\begin{itemize}
\item Section~\ref{sec:novel} presents the novelty and contribution of our approach, providing a comprehensive comparison with other baseline methods.
\item Section~\ref{sec:additional} demonstrates extensive additional experimental results across multiple dimensions. 
\item Section~\ref{sec:setting} details the training settings and hyperparameters for both GraftModel and GraftLoRA implementations.
\item Section~\ref{sec:dataset} provides information on the dataset details used in our experiments.
\item Section~\ref{sec:detail} offers a comprehensive description of the baseline methods employed in our comparative experiments.
\item Section~\ref{sec:limit} discusses the broader impact and limitations of our approach.
\end{itemize}
\section{Novelty and Contribution}\label{sec:novel}
Our research aims to unlock the full potential of parameter fusion approaches by implementing a dual-gate mechanism that addresses parameter competition across different domains. We re-examine existing model merging methods and highlight the critical role of compatibility-aware parameter integration. To clearly demonstrate the innovation of our method, we conduct a comparative analysis with existing state-of-the-art baseline methods.

\noindent \textbf{Comparison with Task Arithmetic.}
Both Task Arithmetic and our Graft approach aim to integrate knowledge from specialized models. However, there are several key differences:
\begin{itemize}
\item Task Arithmetic\cite{ilharco2022editing} employs linear combinations with scalar coefficients, whereas Graft utilizes channel-wise adaptive fusion that captures fine-grained parameter importance.

\item When addressing parameter conflicts, Task Arithmetic\cite{ilharco2022editing} lacks a mechanism to evaluate parameter significance, potentially leading to destructive interference when domains have opposing feature preferences.

\item Our dual-gate fusion mechanism integrates both local channel-level differences and global entropy-based weighting, enabling more nuanced parameter selection.
\end{itemize}
\noindent \textbf{Comparison with TIES-Merging.}
Both TIES-Merging\cite{yadav2023ties}  and our Graft approach address parameter interference through parameter-level adjustments. However, there are several key distinctions:
\begin{itemize}
\item TIES-Merging\cite{yadav2023ties}  employs sign-based pruning with binary decisions, neglecting the continuous spectrum of parameter importance. In contrast, Graft considers the significance of fine-grained parameter by quantifying channel differences.
\item In terms of cross-domain integration, TIES-Merging\cite{yadav2023ties}  only considers parameter sign concordance, whereas our method leverages both functional attribution and information-theoretic signals to guide fusion.

\item Our compatibility scoring mechanism quantifies inter-expert alignment at the activation level, providing a principled approach to domain pair selection that TIES-Merging lacks.
\end{itemize}
\noindent \textbf{Comparison with DARE.}
Both DARE and Graft integrate capabilities from specialized models, but there are significant differences:
\begin{itemize}
\item DARE\cite{yu2024language} applies a mask function to parameter differences without considering domain compatibility, while we employ compatibility analysis to guide parameter fusion.

\item DARE\cite{yu2024language} primarily focuses on preserving the performance of individual experts, whereas Graft specifically targets synergistic integration that enhances cross-domain capabilities.

\item Our activation-based compatibility metric provides a quantitative basis for model selection prior to fusion, which DARE\cite{yu2024language} lacks. This enables more informed fusion decisions, as evidenced by the strong correlation between compatibility scores and performance gains.
\end{itemize}

\section{Addtional Results}\label{sec:additional}
\subsection{Extended Evaluation on other Models}
To further validate the effectiveness and scalability of our approach, we conduct additional experiments on Qwen2.5-VL-3B\cite{bai2025qwen2} across four specialized domains. Table~\ref{tab:qwen25} presents the performance comparison between single-domain models and their fused counterparts across multiple benchmarks. The Math-specialized model achieves the highest MathVista\cite{lu2024mathvistaevaluatingmathematicalreasoning} score of 59.4, while the Code-specialized model excels in HumanEval with 29.3. Our fusion strategy demonstrates consistent improvements: the Math\&Code combination achieves 59.6 on MathVista (+0.2 improvement) while maintaining competitive HumanEval performance at 27.6.

\begin{table}[h]
\begin{center}
\captionsetup{font={small,stretch=1.25}, labelfont={bf}}
  \caption{Performance of Qwen2.5-VL-3B on single‑domain and fused models across multiple domains.
           (\checkmark\ indicates the domain(s) included in the model).}
  \label{tab:qwen25}
  \centering
  \setlength{\tabcolsep}{6pt}
  \renewcommand{\arraystretch}{1.2}

  \resizebox{0.8\textwidth}{!}{%
  \begin{tabular}{cccc||cccc}
    \toprule[2pt]
    \multicolumn{4}{c||}{\textbf{Domain Composition}} &
    \multicolumn{2}{c}{\textbf{Benchmark Scores}} \\
    \hline
    \hline
    \textbf{Math} & \textbf{Code} & \textbf{Finance} & \textbf{Medical} &
    \textbf{MathVista} & \textbf{HumanEval}\\
    \hline

    \checkmark &            &            &            & 59.4 & 25.6\\
               & \checkmark &            &            & 58.2 & 29.3\\
               &            & \checkmark &            & 57.2 & 23.2\\
               &            &            & \checkmark & 57.9 & 12.8\\
    \hline
    \hline
    \checkmark & \checkmark &            &            & 59.6 & 27.6\\
    \checkmark &            & \checkmark &            & 57.8 &  -- \\
    \checkmark &            &            & \checkmark & 57.9 & --  \\
               & \checkmark & \checkmark &            & --   & 29.3\\
               & \checkmark &            & \checkmark & --   & 29.3\\
    \bottomrule[2pt]
  \end{tabular}
  }%
  \end{center}
\end{table}
\subsection{Sample Size Selection Analysis}
\begin{figure*}
    \centering
    \caption{Performance Analysis Across Training Sample Sizes on GraftLoRA.}
    \includegraphics[width=0.8\textwidth]{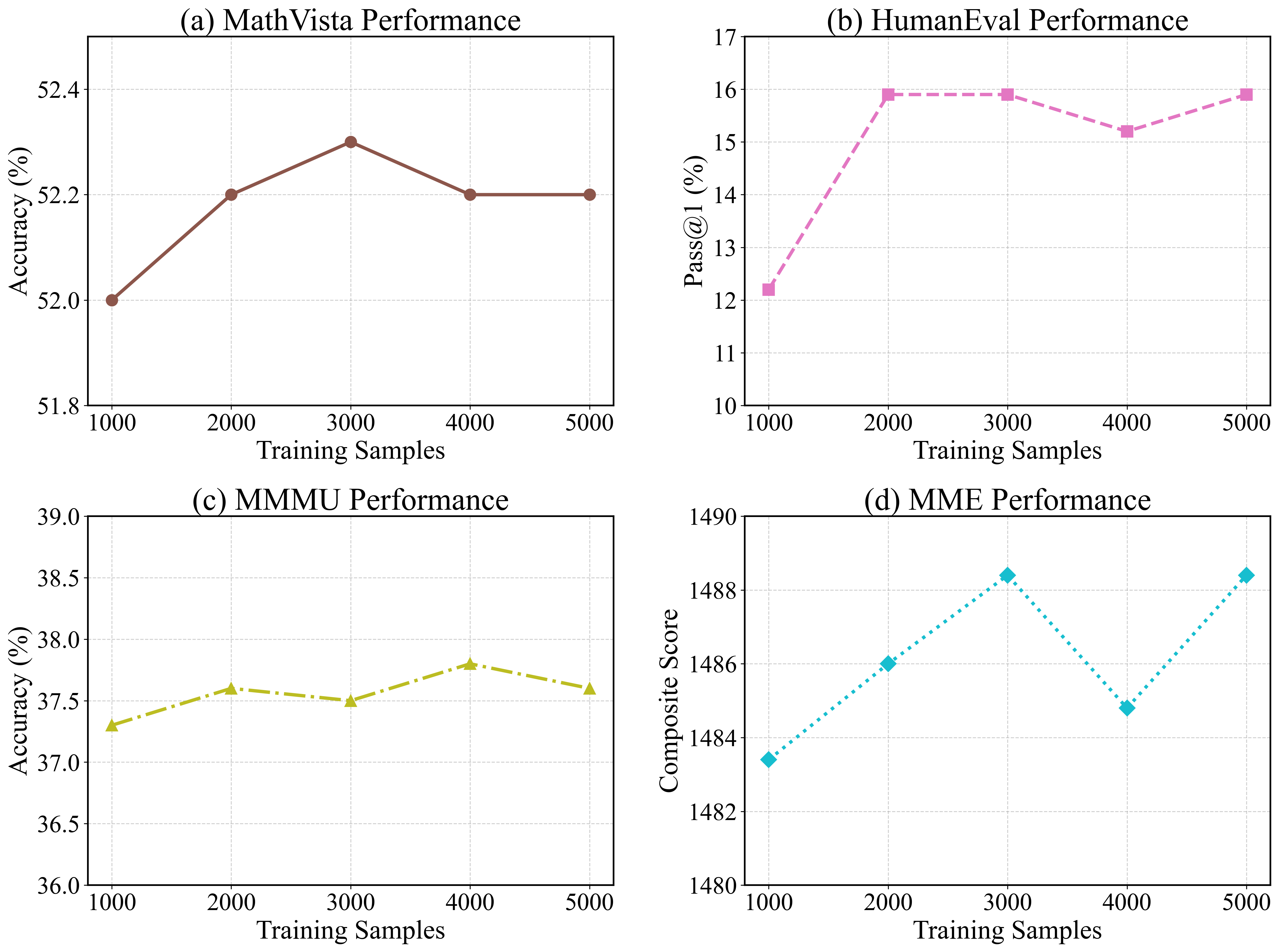}
    
    \vspace{-3mm}
    \label{fig:sample1}
\end{figure*}

\begin{figure*}
    \centering
    \caption{Performance Analysis Across Training Sample Sizes on GraftModel.}
    \includegraphics[width=0.8\textwidth]{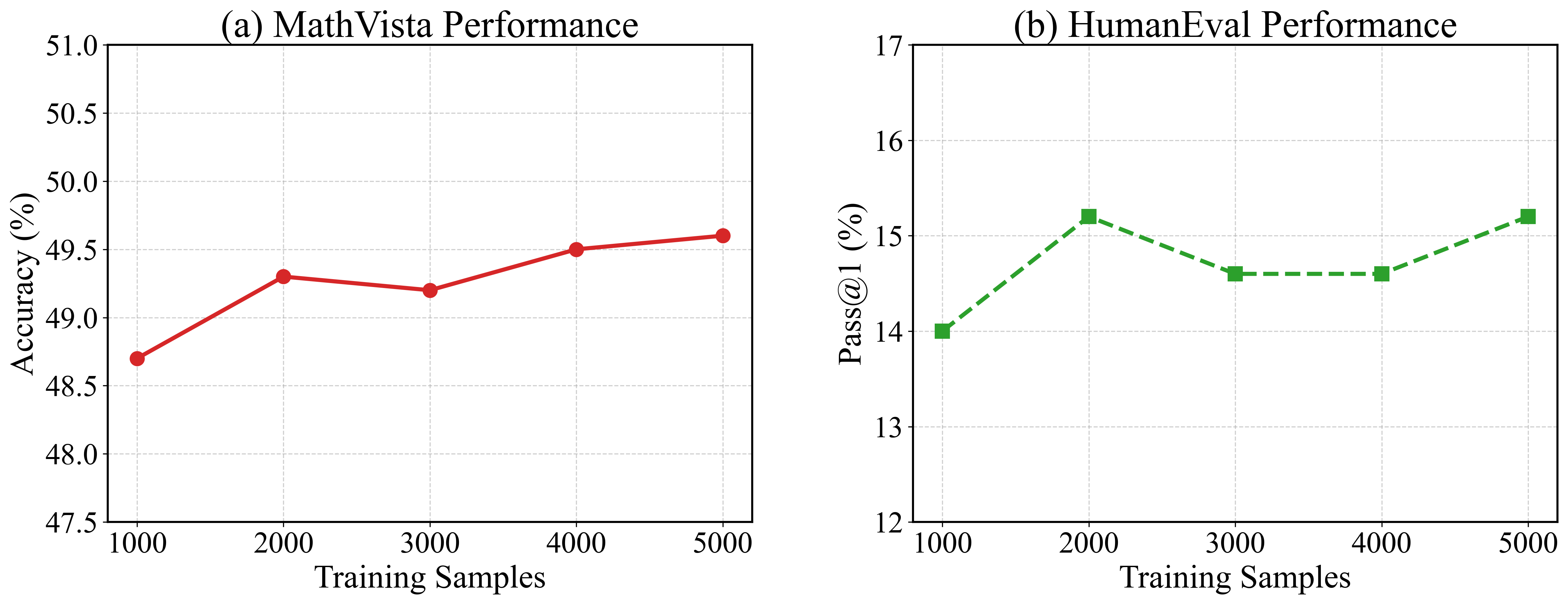}
    
    \vspace{-3mm}
    \label{fig:sample2}
\end{figure*}

To determine the optimal number of samples for our fusion procedure, we conduct a systematic empirical study across varying sample sizes from 1,000 to 5,000. As illustrated in Figure~\ref{fig:sample1} and Figure~\ref{fig:sample2}, our systematic evaluation across varying sample sizes reveals distinct convergence patterns for different fusion approaches. For GraftLoRA fusion, the performance trajectory shows that while peak performance on individual benchmarks occurs at intermediate sample sizes (e.g., 3000 samples achieving 52.3\% on MathVista), the 5000-sample configuration delivers the most consistent performance across all evaluation metrics, achieving 52.2\% on MathVista, 15.9\% on HumanEval, 37.6\% on MMMU, and 1488.4 on MME.

Similarly, GraftModel demonstrates monotonic performance improvements with increasing sample sizes, reaching optimal performance at 5000 samples with 49.6\% on MathVista and 15.2\% on HumanEval. This convergence behavior indicates that our selection of sample size provides sufficient statistical representation for effective parameter alignment while avoiding potential overfitting observed in smaller sample regimes.
\subsection{Fusion Granularity Analysis: Block-wise vs. Channel-wise Parameter Integration}
To comprehensively evaluate the impact of fusion granularity on cross-domain performance, we conduct comparative experiments between our proposed channel-wise fusion strategy and an alternative block-wise approach. The block-wise method operates at an intermediate granularity level, partitioning weight matrices into $8\times8$ blocks, and applying fusion decisions at the block level rather than individual parameters or channels.

\noindent \textbf{Implementation Details.} The block-wise fusion strategy divides each weight matrix $\mathbf{W} \in\mathbb{R}^{M\times N}$ into non-overlapping $8\times8$ blocks. For each block $\mathbf{W}_{block}^{(i,j)}$, we compute block-level differences and apply the dual-gate mechanism to determine fusion weights at the block granularity.  This approach provides a middle ground between parameter-level and channel-level fusion, potentially capturing local parameter correlations while maintaining computational efficiency.

\noindent \textbf{Experimental Results.} Table~\ref{tab:block} presents the performance comparison between single-domain models and their block-wise fused counterparts across multiple benchmarks. The results reveal distinct granularity-dependent performance patterns:

\noindent \textbf{Granularity-Performance Analysis.} The differential performance patterns suggest that fusion granularity exhibits task-dependent optimization characteristics:
\begin{itemize}
    \item \textbf{Mathematical Reasoning Tasks:} It consistently benefits from block-wise fusion, suggesting mathematical reasoning leverages local parameter structures.
    \item \textbf{Code Generation Tasks:} It shows preference for channel-wise precision across all fusion combinations.
    \item \textbf{Multimodal Understanding:} Both approaches achieve almost equivalent scores, showing robustness to moderate granularity variations.
\end{itemize}

\noindent \textbf{Implications.} The systematic analysis reveals that fusion granularity sensitivity correlates with domain-specific computational patterns. Mathematical reasoning benefits from structured parameter blocks, while programming tasks require fine-grained control. Our channel-wise approach provides optimal average performance across diverse domain combinations, validating its adoption as the primary fusion strategy.

\begin{table}[h]
\begin{center}
\captionsetup{font={small,stretch=1.25}, labelfont={bf}}
  \caption{Performance of single‑domain and block-wise fused models across multiple domains.
           (\checkmark\ indicates the domain(s) included in the model).}
  \label{tab:block}
  \centering
  \setlength{\tabcolsep}{6pt}
  \renewcommand{\arraystretch}{1.2}

  \resizebox{0.8\textwidth}{!}{%
  \begin{tabular}{cccc||cccc}
    \toprule[2pt]
    \multicolumn{4}{c||}{\textbf{Domain Composition}} &
    \multicolumn{4}{c}{\textbf{Benchmark Scores}} \\
    \hline
    \textbf{Math} & \textbf{Code} & \textbf{Finance} & \textbf{Medical} &
    \textbf{MathVista} & \textbf{HumanEval} & \textbf{MMMU} & \textbf{MME} \\
    \hline

    \checkmark &            &            &            & 49.9 &  4.3 & 35.6 & 1455.7\\
               & \checkmark &            &            & 47.6 & 15.9 & 37.8 & 1373.9\\
               &            & \checkmark &            & 43.8 &  8.5 & 36.7 & 1414.7\\
               &            &            & \checkmark & 46.0 & 12.2 & 37.5 & 1470.2\\
    \hline
    \hline
    \checkmark & \checkmark &            &            & 53.0 & 14.6 &37.6& 1486.0\\
    \checkmark &            & \checkmark &            & 50.5 &  --  &37.2& 1474.8\\
    \checkmark &            &            & \checkmark & 52.9 & --   &37.3& 1524.3\\
               & \checkmark & \checkmark &            & --   & 14.6 &38.1& 1441.4\\
               & \checkmark &            & \checkmark & --   & 15.9 &37.9& 1470.5\\
    \bottomrule[2pt]
  \end{tabular}
  }%
  \end{center}
\end{table}

\subsection{Additional Evaluation on Pure Language Models}
To demonstrate the versatility and scalability of our GraftLoRA approach beyond multimodal architectures, we conduct supplementary experiments on pure language models. Specifically, we employ Qwen2-1.5B\cite{yang2024qwen2} as the base model and fine-tune two domain-specific experts: a mathematics expert trained on MetaMathQA-40K and a coding expert trained on Code-Alpaca-20K. The resulting single-domain and fused models are evaluated on three established benchmarks: GSM8K for mathematical reasoning, HumanEval for code generation, and HumanEval+ for enhanced code evaluation.

Table~\ref{tab:nlp} presents the comparative performance across these benchmarks. The mathematics expert achieves superior performance on GSM8K\cite{cobbe2021training} (53.6), while the coding expert excels on both programming benchmarks (42.7 on HumanEval\cite{chen2021evaluating}, 39.0 on HumanEval+\cite{liu2023your}). Notably, the GraftLoRA-fused model demonstrates effective cross-domain knowledge integration, achieving 43.6 on GSM8K while maintaining competitive coding performance (42.1 on HumanEval, 40.2 on HumanEval+). Although the fused model exhibits some performance trade-offs in the mathematics domain compared to the specialized expert, it successfully preserves coding capabilities without significant degradation.
\begin{table}[h]
\begin{center}
\captionsetup{font={small,stretch=1.25}, labelfont={bf}}
  \caption{Performance of Qwen2-1.5B on single‑domain and fused models across multiple domains.
           (\checkmark\ indicates the domain(s) included in the model).}
  \label{tab:nlp}
  \centering
  \setlength{\tabcolsep}{6pt}
  \renewcommand{\arraystretch}{1.2}

  \resizebox{0.8\textwidth}{!}{%
  \begin{tabular}{cc||ccc}
    \toprule[2pt]
    \multicolumn{2}{c||}{\textbf{Domain Composition}} &
    \multicolumn{3}{c}{\textbf{Benchmark Scores}} \\
    \hline
    \hline
    \textbf{Math} & \textbf{Code} &
    \textbf{GSM8K} & \textbf{HumanEval} & \textbf{HumanEval+}\\
    \hline

    \checkmark &             &53.6&42.0&38.4\\
               & \checkmark  &41.6&42.7&39.0\\
    \checkmark & \checkmark  &43.6&42.1&40.2\\
    
    \bottomrule[2pt]
  \end{tabular}
  }%
  \end{center}
\end{table}

\subsection{Subtask-Level Further Evaluation.}
Tables~\ref{tab:6} and~\ref{tab:7} provide comprehensive subtask-level analysis that elucidates the granular effectiveness of our fusion approach across diverse cognitive and domain-specific capabilities. The MME benchmark \cite{ref:MME}decomposition (Table~\ref{tab:6}) reveals distinct patterns of cross-domain synergy, with Math\&Medical fusion achieving superior performance in reasoning-intensive tasks (433.6 vs. 400.0 for pure Math) and code reasoning (110.0 vs. 85.0 for pure Code), demonstrating the complementary nature of mathematical and medical domain expertise in complex visual reasoning scenarios.

The subtask analysis reveals task-specific fusion benefits that align with cognitive task requirements. Visual reasoning tasks such as artwork interpretation (144.0 for Math\&Code vs. 126.8 for Math alone) and celebrity recognition (130.3 for Math\&Code vs. 113.2 for Math alone) benefit significantly from cross-domain knowledge integration, while tasks requiring specialized domain knowledge, such as OCR (125.0 for Math) and text translation (177.5 for Math), maintain optimal performance with domain-specific expertise. Notably, certain tasks demonstrate ceiling effects (existence detection achieving 195.0 across all configurations), indicating that our fusion approach preserves maximum performance capabilities without introducing degradation.

The MMMU\cite{yue2024mmmu} domain analysis (Table~\ref{tab:7}) confirms the effectiveness of our fusion methodology across academic disciplines. Code\&Finance fusion excels in Humanities \& Social Science (55.0 vs. 54.4 for Code alone) and Health \& Medicine (39.6 vs. 38.2 for Finance alone), while Code\&Medical fusion demonstrates consistent improvements in Science (32.5 vs. 31.9 for Code alone) and balanced performance across multiple domains. The minimal performance variations (±1.0 point maximum) across domain combinations validate the stability of our dual-gate mechanism in preventing catastrophic interference while enabling selective knowledge transfer. These fine-grained results substantiate that our parameter fusion strategy operates effectively at the cognitive subtask level, preserving specialized capabilities while enhancing cross-domain reasoning through principled parameter integration.
\begin{table}[h]
\captionsetup{font={small,stretch=1.25}, labelfont={bf}}
  \caption{Results of MME benchmark across tasks and domains.}
  \label{tab:6}
  \centering
  \renewcommand{\arraystretch}{1.2}
  \resizebox{0.8\textwidth}{!}{%
  \begin{tabular}{l||c||cc||cc}
    \toprule[2pt]
    \textbf{Model}&\textbf{math}&\textbf{code}&\textbf{math\&code}&\textbf{medical}&\textbf{math\&medical}\\
    \hline
    \hline
    reasoning&400.0&357.1&425.0&372.1&433.6\\
    OCR&125.0&57.5&80.0&95.0&117.5\\
    artwork&126.8&115.0&144.0&138.0&143.25\\
    celebrity&113.2&105.6&130.3&129.7&126.5\\
    code reasoning&75.0&85.0&97.5&77.5&110.0\\
    color&165.0&170.0&175.0&170.0&175.0\\
    commonsense reasoning&105.0&97.1&105.0&82.1&108.6\\
    count&118.3&115.0&130.0&120.0&135.0\\
    existence&195.0&195.0&195.0&195.0&195.0\\
    landmark&160.8&166.8&168.75&169.3&170.5\\
    numerical calculation&42.5&50.0&52.5&50.0&52.5\\
    position&150.0&128.3&148.3&143.3&148.3\\
    posters&150.7&159.5&159.5&143.2&165.0\\
    scene&151.0&161.25&157.5&166.8&159.0\\
    text translation&177.5&125.0&170.0&162.5&162.5\\
    \rowcolor{blue!5}Overall&1455.7&1373.9&1488.4&1470.2&1535.0\\
  \bottomrule[2pt]
  \end{tabular}}
\end{table}

\begin{table}[h]
\captionsetup{font={small,stretch=1.25}, labelfont={bf}}
  \caption{Results of MMMU benchmark across tasks and domains.}
  \label{tab:7}
  \centering
  \renewcommand{\arraystretch}{1.2}
  \resizebox{0.8\textwidth}{!}{%
  \begin{tabular}{l||c||cc||cc}
    \toprule[2pt]
    \textbf{Model}&\textbf{code}&\textbf{finance}&\textbf{code\&finance}&\textbf{medical}&\textbf{code\&medical}\\
    \hline
    \hline
    Art \& Design&53.6&51.4&54.6&53.3&53.7\\
    Business&31.4&30.3&31.4&29.6&30.5\\
    Science&31.9&31.9&31.6&31.4&32.5\\
    Health \& Medicine&38.0&38.2&39.6&38.6&38.6\\
    Humanities \& Social Science&54.4&51.2&55.0&52.7&54.7\\
    Tech \& Engineering&34.1&32.3&33.5&34.6&34.2\\
    \rowcolor{blue!5}Overall&37.8&36.7&38.1&37.5&38.0\\

  \bottomrule[2pt]
  \end{tabular}}
\end{table}

\section{Training Settings}\label{sec:setting}
\subsection{Computational Resources}
All experiments were conducted on Nvidia A6000 GPUs with 48GB of RAM. Depending on the dataset type and size, as well as the fine-tuning type, fine-tuning the Qwen2-VL-2B model on single tasks took between 80 minutes and 75 hours.

Graft experiments took less resources. The fusion process typically completed within 1-4 hours on average, representing a considerable reduction in training time while achieving superior cross-domain performance.

\subsection{Hyper-parameter settings}
The training of our GraftModel method involves several key hyper-parameters and settings to ensure effective fusion of the base and graft Qwen2-VL-2B\cite{wang2024qwen2} models. The specific details are summarized in Table~\ref{tab:hypergraftmodel}. We set the learning rate to $1.0\times10^{-5}$ and trained for a single epoch to prevent overfitting to domain-specific patterns. The training utilized a batch size of 1 per device with gradient accumulation steps of 2, resulting in an effective batch size of 2. We implemented cosine learning rate scheduling with a warmup ratio of 0.1 to ensure smooth convergence during the fusion process. Training was conducted in bfloat16 precision with a maximum sequence length of 2048 tokens, utilizing 16 preprocessing workers for efficient data handling.
\begin{table}[h!]
\centering
\captionsetup{font={small,stretch=1.25}, labelfont={bf}}
\renewcommand{\arraystretch}{1.2}
\caption{Training Hyper-parameters for GraftModel}
\label{tab:hypergraftmodel}
\resizebox{0.48\textwidth}{!}{%
\begin{tabular}{c||c}
\toprule[2pt]
\textbf{Hyper-parameter} & \textbf{Value} \\
\hline
\hline
Finetuning Type & full \\
Maximum Sequence Length & 2048 tokens \\
Preprocessing Workers & 16 \\
Batch Size (per device) & 1 \\
Gradient Accumulation Steps & 2 \\
Effective Batch Size & 2 \\
Learning Rate & 1.0 $\times$ 10$^{-5}$ \\
Training Epochs & 1.0 \\
Learning Rate Schedule & cosine \\
Warmup Ratio & 0.1 \\
Precision & BF16 \\
\bottomrule[2pt]
\end{tabular}%
}
\end{table}

For the GraftLoRA fusion method, it targets both attention and MLP layers with LoRA adaptations, requiring a higher learning rate of $1.0\times10^{-4}$ and extended training over 3 epochs to achieve optimal adapter integration as shown in Table~\ref{tab:hypergraftlora}. 

\begin{table}[h!]
\centering
\captionsetup{font={small,stretch=1.25}, labelfont={bf}}
\caption{Training Hyper-parameters for GraftLoRA}
\renewcommand{\arraystretch}{1.2}
\label{tab:hypergraftlora}
\resizebox{0.48\textwidth}{!}{%
\begin{tabular}{c||c}
\toprule[2pt]
\textbf{Hyper-parameter} & \textbf{Value} \\
\hline
\hline
LoRA Target & attention, mlp \\
Maximum Sequence Length & 2048 tokens \\
Preprocessing Workers & 16 \\
Batch Size (per device) & 4 \\
Gradient Accumulation Steps & 8 \\
Effective Batch Size & 32 \\
Learning Rate & 1.0 $\times$ 10$^{-4}$ \\
Training Epochs & 3.0 \\
Learning Rate Schedule & cosine \\
Warmup Ratio & 0.1 \\
Precision & BF16 \\
\bottomrule[2pt]
\end{tabular}%
}
\end{table}
Both configurations utilize cosine learning rate scheduling with 10\% warmup ratio and BF16 precision to optimize training stability and computational efficiency. The distinct hyper-parameter settings reflect the fundamental differences between full parameter fusion and adapter-based integration, with GraftLoRA requiring more aggressive optimization due to the constrained parameter space of low-rank adaptations.

\section{Dataset Details}\label{sec:dataset}
This section provides comprehensive information about the training datasets and evaluation benchmarks employed in our experiments. Our evaluation framework encompasses both domain-specific fine-tuning datasets and comprehensive multimodal benchmarks to validate cross-domain fusion effectiveness.

\subsection{Training Datasets}
Table~\ref{tab:train1} summarizes the domain-specific datasets used for fine-tuning specialized models. The datasets span four critical domains: mathematics, programming, finance, and medical imaging, providing diverse multimodal reasoning challenges.

\begin{table}[h]
    \centering
    \captionsetup{font={small,stretch=1.25}, labelfont={bf}}
    \caption{Detailed Description of Training Datasets}
    \renewcommand{\arraystretch}{1.2}
    \label{tab:train1}
    \begin{tabular}{c||c||c||c}
        \toprule[2pt]
        \textbf{Dataset}& \textbf{Domain} & \textbf{\#Train} & \textbf{Modality} \\ 
        \hline
        \hline
        
        MathV-360K\cite{shi2024math}&Mathematics&338,721&Vision+Text \\
        Code-Alpaca-20K\cite{codealpaca}&Programming&20,021&Text \\
        Sujet-Finance-QA-Vision-100K\cite{Sujet-Finance-QA-Vision-100k}&Finance&100,629&Vision+Text \\
        PathVQA\cite{he2020pathvqa30000questionsmedical}&Medical&19,654&Vision+Text \\
        MetaMathQA-40K\cite{yu2023metamath}&Mathematics&40,000&Text \\
        
        \bottomrule[2pt]
    \end{tabular}
\end{table}

\subsection{Data Format}
Our training datasets adopt two distinct organizational structures in order to meet different modality requirements and task characteristics.

\noindent \textbf{Multimodal Question Format.} For vision-language tasks (MathV-360K\cite{shi2024math}, Sujet-Finance-QA-Vision-100K\cite{Sujet-Finance-QA-Vision-100k}, PathVQA\cite{he2020pathvqa30000questionsmedical}), data follows a conversational structure with two primary components:

\begin{itemize}
    \item \textbf{Message Exchange:} User-assistant dialogue pairs where user messages contain visual references ("<image>") alongside textual queries, and assistant messages provide domain-specific responses.
    \item \textbf{Image References:} Explicit file paths linking visual content to the corresponding textual interactions.
\end{itemize}

\noindent \textbf{Instruction-Response Format.} For text-only tasks (Code-Alpaca-20K\cite{codealpaca}, MetaMathQA-40K\cite{yu2023metamath}), data employs a structured triplet organization:

\begin{itemize}
    \item \textbf{Instruction:} Task specifications and objectives that define the expected behavior or solution approach.
    \item \textbf{Input:} Additional context or parameters required for task completion (may be empty for self-contained instructions).
    \item \textbf{Output:} Target responses demonstrating correct task execution, including code implementations or mathematical solutions.
\end{itemize}

\subsection{Evaluation Benchmarks}
Table~\ref{tab:evaluation} details the comprehensive evaluation benchmarks used to assess both single-domain expertise and cross-domain fusion effectiveness. These benchmarks cover diverse cognitive capabilities including mathematical reasoning, code generation, multimodal understanding, and domain-specific knowledge assessment.
\begin{table}[h]
    \centering
    \captionsetup{font={small,stretch=1.25}, labelfont={bf}}
    \caption{Detailed Description of Evaluation Benchmarks}
    \renewcommand{\arraystretch}{1.2}
    \label{tab:evaluation}

    \begin{tabular}{c||c||c||c}
        \toprule[2pt]
        \textbf{Benchmark}& \textbf{Domain} & \textbf{\#Test} & \textbf{Metrics} \\ 
        \hline
        \hline
        
        MathVista\cite{lu2024mathvistaevaluatingmathematicalreasoning}&Mathematics&12 subtasks&Accuracy (\%) \\
        GSM8K\cite{cobbe2021training}&Mathematics&1319 problems&Accuracy (\%)\\
        HumanEval\cite{chen2021evaluating}&Programming&164 problems&Pass@1 (\%) \\
        HumanEval+\cite{liu2023your}&Programming&164 problems&Pass@1 (\%) \\
        MMMU\cite{yue2024mmmu}&Multimodal&6 disciplines&Accuracy (\%) \\
        MME\cite{ref:MME}&Multimodal&14 subtasks&Composite Score \\
        
        \bottomrule[2pt]
    \end{tabular}
\end{table}

\section{Baseline Details}\label{sec:detail}
This section provides a comprehensive description of the baseline methods employed in our comparative experiments. We evaluate our approach against the following established methods:

\textbf{Parameter-Based Merging Methods}
\begin{itemize}
    \item \textbf{Weight Averaging:} The most straightforward model merging technique that computes the element-wise mean of parameters from all individual models: $\theta_m=\sum^{n}_{t=1}\theta_t/n$.
\end{itemize}

\textbf{Vector-Based Merging Methods}
\begin{itemize}
    \item \textbf{Task Arithmetic:} This method introduces the concept of "task vectors" and merges these vectors into a pre-trained model to enable multi-task capability. The merged model is formulated as: $\theta_m=\theta_{init}+\lambda\cdot\sum^{n}_{t=1}\tau_t$, where $\theta_{\text{init}}$ is the initial pre-trained model, $\tau_t$ represents task-specific vectors, and $\lambda$ controls the merging intensity.

    \item \textbf{Ties-Merging:} This approach addresses task conflicts present in Task Arithmetic through a three-step process: Trim (eliminating redundant parameters), Elect Sign (resolving directional conflicts), and Disjoint Merge (combining non-conflicting parameters). This methodical approach enhances multi-task performance by reducing parameter interference.
\end{itemize}

\textbf{Module-Based Merging Methods}
\begin{itemize}
    \item \textbf{DARE:} This method enhances parameter efficiency by setting the majority of delta parameters to zero and rescaling the remaining parameters: $\theta' = \theta\cdot\frac{1}{1-p}$, where $p$ represents the proportion of delta parameters eliminated. This approach effectively reduces parameter redundancy while maintaining performance.
\end{itemize}

\section{Boarder Impact and Limitations}\label{sec:limit}
\noindent \textbf{Boarder Impact.} The Graft framework advances parameter synergy for Multimodal Large Language Models (MLLMs) by enabling efficient domain knowledge integration without extensive retraining. This approach significantly reduces computational demands for developing domain-adaptive AI systems, offering both economic and environmental benefits through more efficient resource utilization. By enabling modular composition of specialized capabilities, Graft democratizes access to high-performance multimodal systems across diverse application domains.

However, enhanced MLLM capabilities raise important considerations. The integrated models may inherit or potentially amplify biases present in constituent domain experts. Additionally, improved performance across multiple domains may lower barriers to generating synthetic content requiring careful verification. As with all advanced AI technologies, responsible deployment practices remain essential as these systems evolve and proliferate.

\noindent \textbf{Limitations.} While Graft demonstrates effective parameter integration, several limitations represent opportunities for future research:
\begin{itemize}
    \item \textbf{Hyperparameter Sensitivity:} The global gating adjustment parameters (a and c) were set empirically based on preliminary experiments. Although performance remained stable across a reasonable range of values, an automated hyperparameter selection process could further optimize fusion for specific domain combinations.

    \item \textbf{Documentation Requirements:} Effective use of our compatibility analysis requires maintaining proper documentation of domain-specific models and their intended applications. Without such documentation, users might attempt to merge incompatible domains, potentially leading to suboptimal results despite technically successful fusion.
\end{itemize}

\newpage
\bibliographystyle{unsrt}
\bibliography{ref}